\documentclass[runningheads,a4paper]{llncs}

\usepackage{amssymb}
\usepackage{graphicx}

\usepackage{verbatim}

\begin{document}

\mainmatter  

\title{Personalized Pancreatic Tumor Growth Prediction via Group Learning}

\titlerunning{Personalized Pancreatic Tumor Growth Prediction}

\author{Ling Zhang\footnote[1]{ling.zhang3@nih.gov}\inst{1} \and Le Lu\inst{1} \and Ronald M. Summers\inst{1} \and Electron Kebebew\inst{2} \and \\ Jianhua Yao\inst{1}}

\institute{
$^1$ Imaging Biomarkers and Computer-Aided Diagnosis Laboratory and the Clinical Image Processing Service, Radiology and Imaging Sciences Department, National Institutes of Health Clinical Center, Bethesda, MD 20892, USA \\
$^2$ Endocrine Oncology Branch, National Cancer Institute, National Institutes of Health, Bethesda, MD 20892, USA
}

\toctitle{Lecture Notes in Computer Science}
\tocauthor{Authors' Instructions}
\maketitle

\begin{abstract}
Tumor growth prediction, a highly challenging task, has long been viewed as a mathematical modeling problem, where the tumor growth pattern is personalized based on imaging and clinical data of a target patient. Though mathematical models yield promising results, their prediction accuracy may be limited by the absence of population trend data and personalized clinical characteristics. In this paper, we propose a statistical group learning approach to predict the tumor growth pattern that incorporates both the population trend and personalized data, in order to discover high-level features from multimodal imaging data. A deep convolutional neural network approach is developed to model the voxel-wise spatio-temporal tumor progression. The deep features are combined with the time intervals and the clinical factors to feed a process of feature selection. Our predictive model is pretrained on a group data set and personalized on the target patient data to estimate the future spatio-temporal progression of the patient's tumor. Multimodal imaging data at multiple time points are used in the learning, personalization and inference stages. Our method achieves a Dice coefficient of $86.8\% \pm 3.6\%$ and RVD of $7.9\% \pm 5.4\%$ on a pancreatic tumor data set, outperforming the DSC of $84.4\% \pm 4.0\%$ and RVD $13.9\% \pm 9.8\%$ obtained by a previous state-of-the-art model-based method.
\end{abstract}

\section{Introduction}
Pancreatic neuroendocrine tumors 
are slow-growing, and usually are not treated until they reach a certain size. 
To choose between nonoperative or surgical treatments, and 
 to better manage the treatment planning, it is crucial to accurately predict the patient-specific spatio-temporal progression of pancreatic tumors \cite{wong2016pancreatic}. 

The prediction of tumor growth is a very challenging task. It has long been viewed as a mathematical modeling problem \cite{clatz2005realistic,hogea2007modeling,wong2016pancreatic}. 
Clinical imaging data provide non-invasive and in vivo measurements 
of the tumor over time at a macroscopic level. For this reason, previous works on image-based tumor growth modeling are mainly based on the reaction-diffusion equations and on biomechanical models. Some previous tumor growth models \cite{clatz2005realistic,hogea2007modeling,wong2016pancreatic} are derived from two or more longitudinal imaging studies of a specific patient over time. While they yield promising results, they fail to account for the population trend of tumor growth patterns and specific tumor clinical characteristics.

Aside from mathematical modeling methods, the combination of data-driven principles and statistical group learning may provide a potential solution to solve these problems by building a model based on both population trend and personalized clinical characteristics. The only pioneer study in this direction \cite{morris2006learning} attempts to model the glioma growth patterns in a classification-based framework. This model learns tumor growth patterns from selected features at the patient-, tumor-, and voxel-levels, and achieves a prediction accuracy of 59.8\%. However, this study only uses population trend of tumor growth without incorporating the history of the patient-specific tumor growth pattern, and is unable to predict tumor growth at different time points. Furthermore, this early study only employs hand-crafted low-level features.  
In fact, information describing tumor progression may potentially lie in the latent high level feature space of tumor imaging, but this has yet to be investigated.  

Representation learning, which automatically learns intricate discriminative information from raw data,  has been popularized by deep learning techniques, namely deep convolutional neural networks (ConvNets) \cite{krizhevsky2012imagenet}. ConvNets have significantly improved quantitative performance on a variety of medical imaging applications \cite{greenspan2016guest}. 
The idea is using deep learning to determine the current status of a pixel or an image patch (whether it belongs to object boundary/region, or a certain category). The ConvNets have been used in prediction of future status of image level - disease outcomes, such as survival prediction of lung cancer patients \cite{yao2016imaging}. However, it is still unknown whether deep ConvNets are capable of predicting the future status at the pixel/voxel level, such as later pixel subsequent involvement regions of a tumor. 

In this paper, we propose a statistical group learning framework to predict tumor growth that incorporates tumor growth patterns derived from population trends and personalized clinical factors. Our hypothesis is that regions involved in future tumor progression is predictable by combining visual interpretations of the longitudinal multimodal imaging information with those from clinical factors. 

\begin{figure}[!t]
   \begin{center}
   \begin{tabular}{c}
   \includegraphics[width=9.5cm]{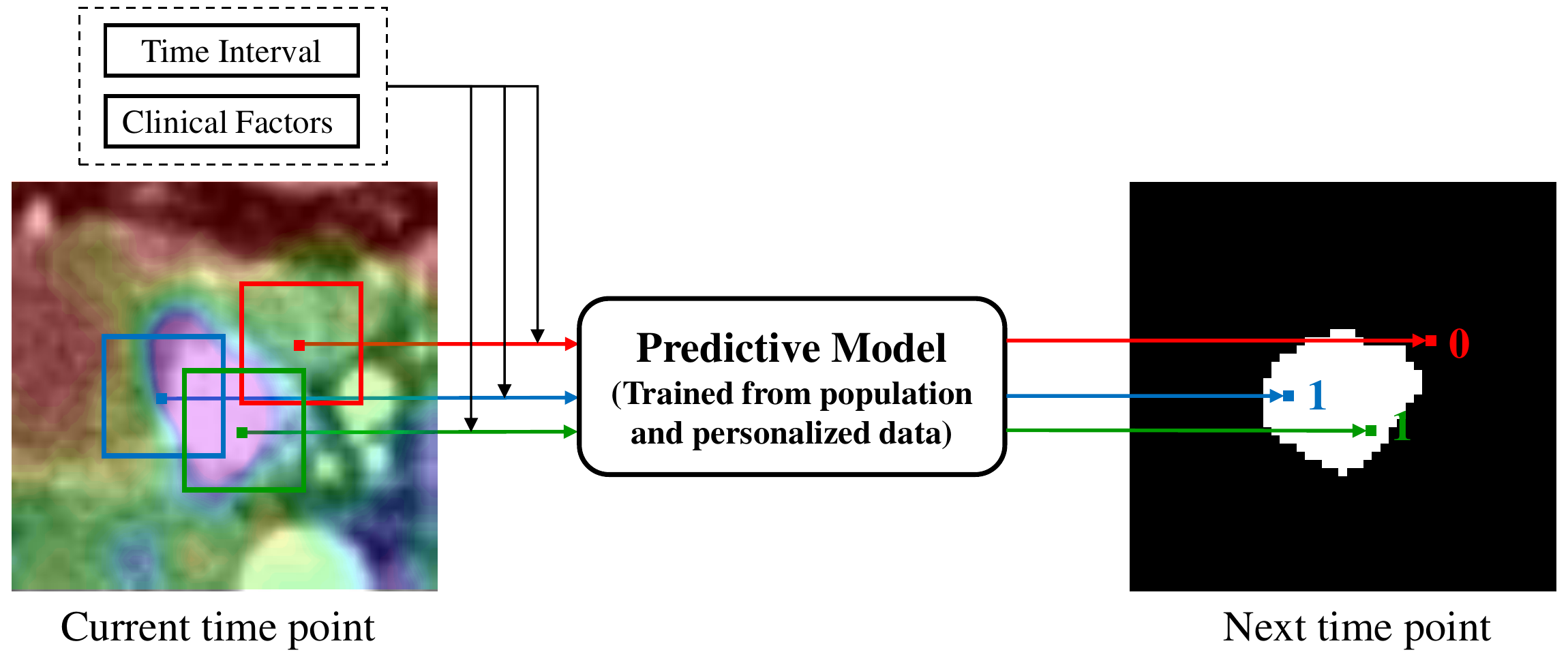}
   \end{tabular}
   \end{center}
   \caption[example] 
   { \label{figidea} 
Framework of the voxel-wise prediction of tumor growth via statistical learning. 
}
   \end{figure}

Our main objective is to design a deep learning predictive model to predict whether the voxels in the current time point will become tumor voxels or not at the next time point (cf. Fig. \ref{figidea}). First, the ConvNet is used to discover the high-level features from multimodal imaging data that carry different aspects of tumor growth related information: (1) FDG-PET (2-[18F] Fluoro-2-deoxyglucose positron emission tomography), to measure the metabolic rate; (2) dual-phase CT, to quantify the physiological parameter of the cell density and to delineate the tumor boundary. 
An example of such multimodal data (color-coded PET overlays on CT) is shown in  Fig. \ref{figidea}. Second, the extracted deep features are combined with time intervals, tumor-level features and clinical factors to form a concatenated feature vector, from which a robust feature subset is selected by the support vector machine recursive feature elimination (SVM RFE) technique \cite{guyon2002gene}, regularized with prior knowledge. Third, a SVM predictive model is trained on a group dataset and personalized on the target patient data to predict the tumor's spatio-temporal growth and progression. 

Our proposed group learning method is compared with a state-of-the-art model-based method \cite{wong2016pancreatic} on a pancreatic tumor growth dataset, and attains both superior accuracy and efficiency. These results highlight the relevance of tumor high-level visual information, as well as tumor- and patient-level features, for predicting the spatio-temporal progression of pancreatic tumors. Our contributions are two-fold: (1) To the best of our knowledge, this is the first adoption of deep ConvNets in voxel-wise prediction of future voxel status, especially to learn the spatio-temporal progression pattern of tumors from multimodal imaging; (2) The proposed method allows for incorporating tumor growth patterns from a group data set and personalized data into a statistical learning framework.

\section{Group Learning Approach for Tumor Growth Prediction}

   \begin{figure}[!t]
   \begin{center}
   \begin{tabular}{c}
   \includegraphics[width=11cm]{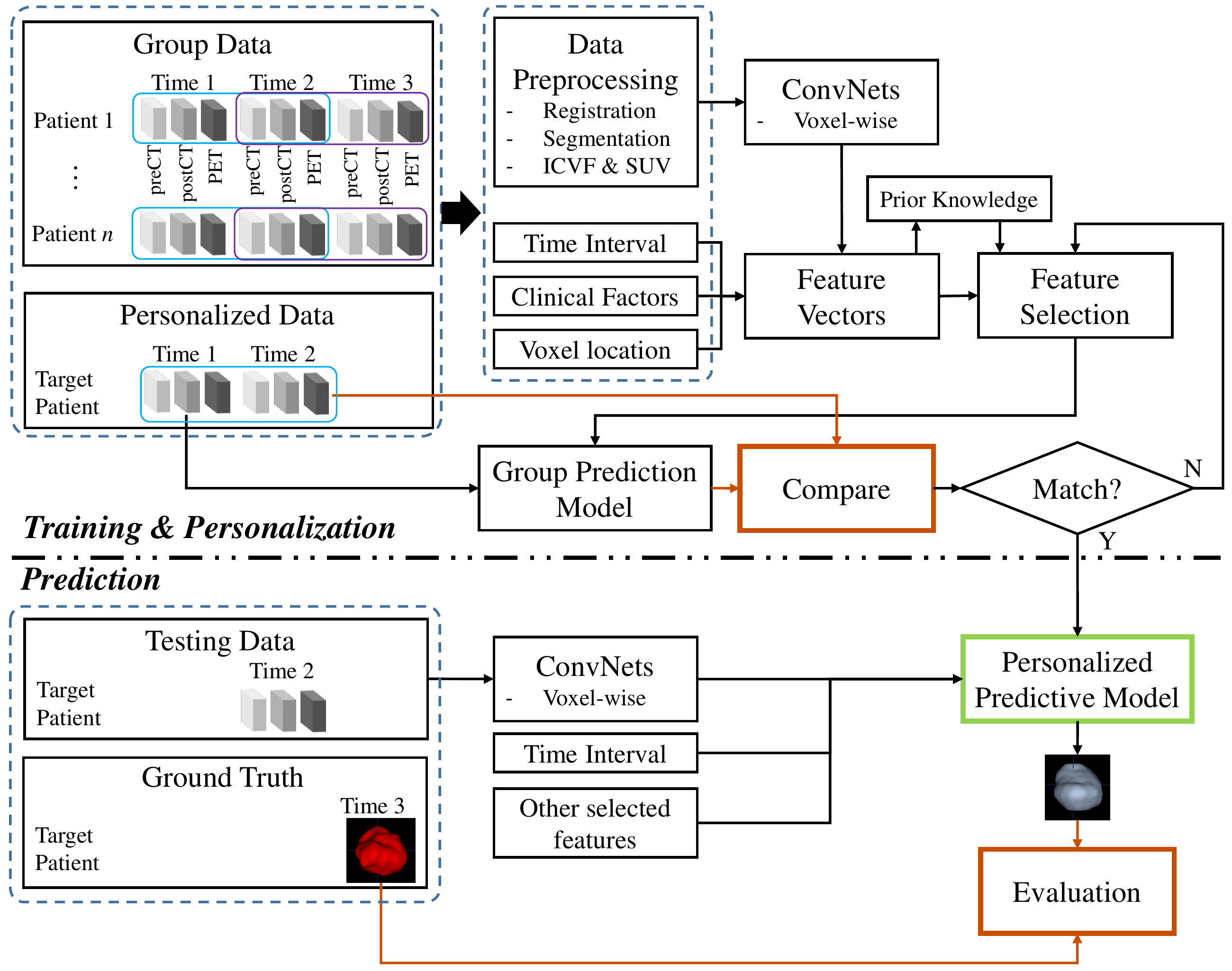}
   \end{tabular}
   \end{center}
   \caption[example] 
   { \label{figoverview} 
Overview of the proposed learning method for predicting tumor growth. The upper part represents stages of model training (to learn population trend) \& personalization and the lower part formulates the process of (unseen) data prediction.
}
   \end{figure} 

In the longitudinal pancreatic tumor data studied in this work, each patient has multimodal imaging data (dual phase contrast-enhanced CT and FDG-PET) and clinical records at three time points spanning $3-4$ years. We design an integrated training \& personalization and prediction framework  illustrated in Fig. \ref{figoverview}. The imaging data scans of different modalities acquired at different time points are first registered, after which the tumors are segmented. Intracellular volume fraction (ICVF) and standardized uptake value (SUV) \cite{wong2016pancreatic} are also computed. In the training \& personalization stage, all voxel-wise ConvNets- and location-based features, time intervals, and clinical factors are extracted from any pairs of two time points (time1/time2 and time2/time3) from group data (patient 1 -- patient $n$) and the pair of time1/time2 from personalized data (the target patient, denoted as patient $n+1$). 
Next, feature selection, which takes prior knowledge into account, is used to rank these features from hybrid resources. The top $m$-ranked features ($m=1,...,M$) are employed to train SVM models on group data (to capture population trend). These SVM classifiers are then personalized via the time1/time2 pair of the target patient data to determine the optimal feature set and model parameters (personalization). In the prediction stage, given the data of the target patient at time2, the imaging and clinical features are fed into the predictive model to predict and estimate the voxel-wise tumor region at a future time3. Note that the testing data (i.e., for predicting time3 based on time2 of the target patient) has never been seen by the predictive model.

\subsection{Image Processing and Patch Extraction}

To establish the spatio-temporal relationship of tumor growth along different time points, the multi-model patient imaging datasets are registered using mutual information, and imaging data at different time points are aligned at the tumor center \cite{wong2016pancreatic}. Afterwards, three types of information related to tumor properties are extracted from the multimodal images and preprocessed as a three-channel image to be fed into ConvNets. Image-specific preprocessing steps include the following: (1) SUV values from PET images are magnified by 100 followed by a cutting window [100 2600] and then linearly transformed to [0 255]; (2) ICVF values are magnified by 100 (range between [0 100]); and (3) tumor mask/boundary is obtained by a level set algorithm \cite{wong2016pancreatic}.

As illustrated in Fig. \ref{figidea}, 
image patches of size $s \times s$ centered at voxels around the tumor region at the current time point are extracted. Patch centers locate inside or outside of the tumor region at the next time point are labelled as ``1"s and ``0"s, respectively, and serve as positive and negative training samples. 
The patch center sampling range is restricted to a bounding box of $\pm 15$ voxels centered at the tumor center, given that the pancreatic tumors in this dataset do not exceed 3 cm ($\approx 30$ voxels) in diameter and are slow growing. 
 To improve the training accuracy and convergence rate of the ConvNet \cite{krizhevsky2012imagenet}, we balance the class distribution of the training set by proportionally under-sampling the non-tumor negative patches. 
The patch-based strategy compensates the small size of longitudinal tumor dataset.

\subsection{Learning a Voxel-Wise Deep Representation}
We use AlexNet \cite{krizhevsky2012imagenet} as our network architecture. AlexNet contains five convolutional ($conv1 - conv5$), three pooling ($pool1$, $pool2$, $pool5$), and two fully connected layers ($fc6 - fc7$). 
 This network is trained from scratch on all pairs of time points (time1/time2 and time2/time3) from the group data set. The training is terminated after a pre-determined number of epochs, where the model with the lowest validation loss is selected as the final network. 

The resulting ConvNet is then used to extract the high-level representation of voxels/patches. This is achieved by feeding the three-channel SUV-ICVF-mask image patches into the personalized ConvNet model, where the $fc$ and the output layers can be treated as the learned deep features. Considering that the high dimensional deep image features of the $fc$ layers may tend to overwhelm the low number tumor- and patient-level features if combined directly, the outputs of the last layer with two nodes are regarded as the final extracted deep features.

\subsection{Learning a Predictive Model with Multi-Source Features}

\subsubsection{Feature Extraction and Selection} A general statistical learning concept is that cues from different sources can provide complementary information for learning a stronger classifier. Therefore, in addition to deep features, we extract three other types of features: (1) Time intervals between two imaging time points, with days as the time unit. (2) Tumor-level features --  the Euclidean distance of the patch center towards its closest tumor surface within the 3D volume for each voxel. This distance value is positive if the patch center locates inside the current tumor region and negative otherwise. In addition, the tumor volume is calculated. (3) Patient-level features, including age, gender, height, and weight.

The SVM RFE technique \cite{guyon2002gene} is adopted to find the most informative features during the process of model training \& personalization. 
 Reflecting the significance of image-based features for assessing the growth of tumor \cite{wong2016pancreatic}, the two deep features are found to be always selected by the SVM RFE model selection. 
Finally, time interval is used as a prior feature, as it is necessary for our task. 

\subsubsection{Predictive Model Training \& Personalization, and Testing}
\label{trainvaltest}
Once the feature set has been fully ranked, the first $m$ features ($m$=[2, 3, ..., 9]) are each iteratively added to train a set of (eight) SVM classifiers until all features are included. In each iteration, the SVM classifier 
is trained on samples from the group data set, and then personalized/validated on the samples of the personalization data set. The validation accuracies are calculated and recorded for all classifiers, where the accuracy metric ($ACC$) is defined by $ACC = \frac{TP+TN}{TP+FP+FN+TN}$. The feature set and classifier that maximize the validation $ACC$ are selected.

To better personalize the predictive model from population trend to the target patient, we optimize an objective function which measures the agreement between the predicted tumor volume and its future ground truth volume on the target patient. To do so, we first apply the predictive model to voxels in the searching neighborhood (tumor growth zone) of the personalization volume, and later threshold the classification outputs. The relative volume difference (RVD) between the predicted and ground truth tumor volumes are computed. As in \cite{wong2016pancreatic}, the tumor growth zone is set as a bounding box surrounding the tumor, parametrized with the pixel distances $N_{x}$, $N_{y}$, and $N_{z}$  to the tumor surface in the $x$, $y$, and $z$ directions, respectively. 

In the testing stage, given the data at time 2 of the target patient, the predictive model, along with its personalized model parameters, is applied to predict the label of every voxel in the growth zone at time 3 .

\section{Experiments and Results}

Seven pancreatic neuroendocrine tumors from seven patients (five males and two female) are studied. These tumors are not treated until they reach 3 cm in diameter, which is the size threshold for treatment for this particular disease. The average age, height and weight of the patients at time 1  were 48.6$\pm$13.9 years, 1.70$\pm$0.13 meters, and 88.1$\pm$16.7 kg respectively. The time interval between two time points is $418 \pm 142$ days (mean $\pm$ std.). This dataset is obtained from \cite{wong2016pancreatic}. 

The ConvNet is trained over 30 epochs. The initial learning rate is 0.001, and is decreased by a factor of 10 at every tenth epoch. Weight decay and momentum are set to 0.0005 and 0.9. A dropout ratio of 0.5 is used to regularize the $fc6$ and $fc7$ layers. Mini-batch size is 256. The image patch size $s$ is set as 17 pixels due to the small size of the pancreatic tumors. To accomodate the Caffe framework used for our ConvNet, the original $17 \times 17$ image patches are up-sampled to $256\times 256$ patches via bi-linear interpolation. 
A total of 36,520 positive and 41,999 negative image patches is extracted from seven patients. AlexNet is run on the Caffe platform \cite{jia2013caffe}, using a  NVIDIA GeForce GTX TITAN Z GPU with 12 GB of memory. The SVM (LIBSVM library \cite{chang2011libsvm}) with linear kernel ($C=1$) is used for both SVM RFE feature selection and SVM classifier training. The parameters for the tumor growth zone are set as $N_{x}=3$, $N_{y}=3$, and $N_{z}=3$ for prediction speed concern, and we note that the prediction accuracy is not sensitive to variation of these parameters.

We evaluate the proposed method using a leave-one-out cross-validation, which not only facilitates comparison with the state-of-the-art model-based method \cite{wong2016pancreatic} (tumor status at time1 and time2 already known, predict time3), but more importantly enables learning both population trend and patient-specific tumor growth patterns. In each of the 7 validations, 6 patients are used as the group training data to learn the population trend, the time1/time2 of the remaining patient is used as the personalization data set, and time2/time3 of the remaining patient as the testing set. We obtain the model's final performance values by averaging results from the 7 cross validation folds. The prediction performance is evaluated using measurements at the third time point by four metrics: recall, precision, Dice coefficient, and RVD (as defined in \cite{wong2016pancreatic}).


 \begin{figure}[!t]
 \begin{center}
 \begin{tabular}{c}
 \includegraphics[width=11cm]{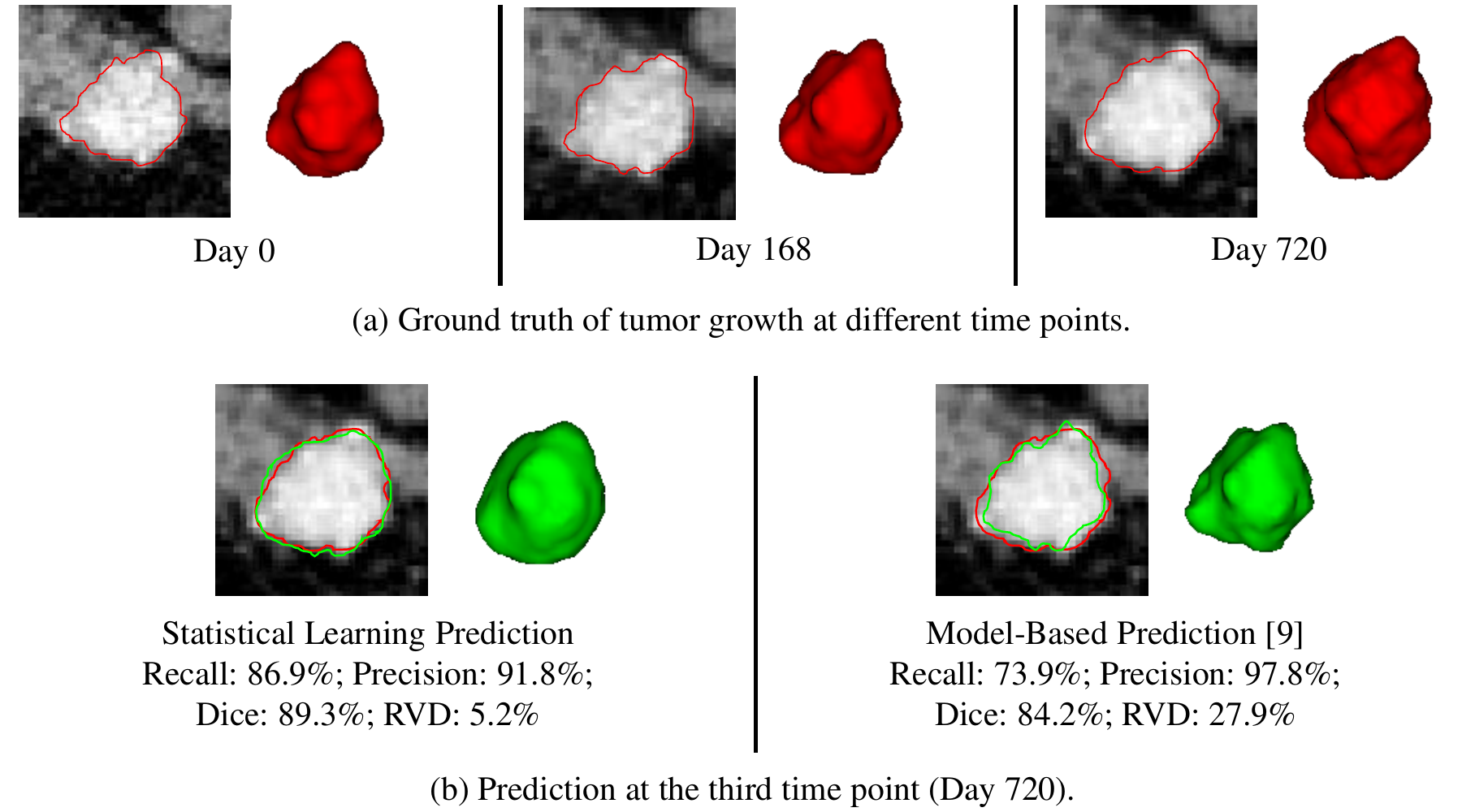}
 \end{tabular}
 \end{center}
 \caption[example] 
 { \label{figresult} 
Comparison of the proposed learning based tumor growth prediction to a state-of-the-art model-based prediction \cite{wong2016pancreatic}. (a) Segmented (ground truth) tumor contours and volumes at different time points. (b) Prediction results at the third time point obtained by learning and model-based techniques (red: ground truth boundaries; green: predicted tumor boundaries).
}
 \end{figure}

\begin{table}[!htbp] 
\scriptsize
\centering
\caption{Performance comparison of our method with the model-based method (EG-IM framework \cite{wong2016pancreatic}) on testing set. Results are reported as: mean $\pm$ std [min, max].
}
\label{performance}
\begin{tabular}{|p{1cm}|p{2.7cm}|p{2.7cm}|p{2.7cm}|p{2.5cm}|}
\hline
 & Recall (\%) & Precision (\%) & Dice (\%) & RVD (\%)\\
\hline
Ref. \cite{wong2016pancreatic} & 83.2$\pm$8.8 [69.4, 91.1] & \textbf{86.9}$\pm$8.3 [74.0, \textbf{97.8}] & 84.4$\pm$4.0 [79.5, \textbf{92.0}] & 13.9$\pm$9.8 [3.6, 25.2] \\
\hline
Ours & \textbf{87.9}$\pm$\textbf{5.0} [\textbf{81.4}, \textbf{94.4}] & 86.0$\pm$\textbf{5.8} [\textbf{78.7}, 94.5] & \textbf{86.8}$\pm$\textbf{3.6} [\textbf{81.8}, 91.3] & \textbf{7.9}$\pm$\textbf{5.4} [\textbf{2.5}, \textbf{19.3}] \\
\hline
\end{tabular}
\end{table}

 In the example shown in Fig. \ref{figresult}, our method achieves both a higher Dice coefficient and a lower RVD than the model-based method. 
Note that the perfect values for Dice and RVD are 100\% and 0\%, respectively.
As indicated in Table \ref{performance}, our method yields a higher Dice coefficient ($86.8 \pm 3.6\%$ vs. $84.4 \pm 4.0\%$), and especially a much lower RVD ($7.9 \pm 5.4\%$ vs. $13.9 \pm 9.8\%$), than the model-based method \cite{wong2016pancreatic}, and thus is far more effective in future tumor volume prediction. The model-based approach in \cite{wong2016pancreatic} requires $\sim$ 24 hrs for model personalization and $\sim 21$ s for simulation per patient, while our method merely requires 3.5 hrs for training and personalization and $4.8 \pm 2.8$ minutes for prediction per patient. 

\section{Conclusion}
In this paper, we have demonstrated that our statistical group learning method, which incorporates tumor growth patterns from a population trend and a specific patient, deep image confidence features, and time interval and clinical factors in a robust predictive model, is an effective approach for tumor growth prediction. Experimental results validate the relevance of tumor high-level visual information coupled tumor- and patient-level features for predicting the spatio-temporal progression of pancreatic tumors. The proposed method outperforms a state-of-the-art model-based method \cite{wong2016pancreatic}. However, it does not consider crucial tumor biomechanical properties, such as tissue biomechanical strain measurements. We plan to include such information in future work, where we will combine deep learning and model-based methods to design  an even more comprehensive and robust predictive model. 


\bibliography{ref}
\bibliographystyle{splncs03}

\end{document}